\RequirePackage{fix-cm}
\documentclass[letter,11pt]{article}

\usepackage[ margin=1in]{geometry}
\usepackage{times}
\usepackage{graphicx}
\usepackage{wrapfig}
\usepackage{subcaption}
\usepackage{setspace}
\usepackage{multicol}
\usepackage{comment} 
\usepackage[T1]{fontenc}
\usepackage[normalem]{ulem}
\usepackage{enumitem}
\usepackage{amsmath}
\usepackage{amssymb}
\usepackage{fancyhdr}
\usepackage{lastpage}
\usepackage{hyperref}
\usepackage{apacite}
\usepackage{xcolor}
\title{On the Compatibility of Generative AI and Generative Linguistics}
\date{May 2025}
\author{\textbf{Eva Portelance} \\
Department of Decision Sciences,\\
HEC Montréal \\
Mila - Quebec Artificial Intelligence Institute \\
\texttt{eva.portelance@hec.ca} \\
\and
\textbf{Masoud Jasbi} \\
Department of Linguistics, \\
University of California - Davis \\
\texttt{jasbi@ucdavis.edu}
}
\begin{document}

\maketitle

\begin{abstract}
In mid-20th century, the linguist Noam Chomsky established \textit{generative linguistics}, and made significant contributions to linguistics, computer science, and cognitive science by developing the computational and philosophical foundations for a theory that defined language as a formal system, instantiated in human minds or artificial machines. These developments in turn ushered a wave of research on symbolic Artificial Intelligence (AI). More recently, a new wave of non-symbolic AI has emerged with neural Language Models (LMs) that exhibit impressive linguistic performance, leading many to question the older approach and wonder about the the compatibility of generative AI and generative linguistics. In this paper, we argue that generative AI is compatible with generative linguistics and reinforces its basic tenets in at least three ways. First, we argue that LMs are formal generative models as intended originally in Chomsky’s work on formal language theory. Second, LMs can help develop a program for discovery procedures as defined by Chomsky's \textit{Syntactic Structures}. Third, LMs can be a major asset for Chomsky’s minimalist approach to Universal Grammar and language acquisition. In turn, generative linguistics can provide the foundation for evaluating and improving LMs as well as other generative computational models of language.
\end{abstract}

\section{Introduction}

Theoretical linguistics has the central goal of describing and explaining our \textit{knowledge of language}, including how language is \textit{acquired}, \textit{understood}, and \textit{produced} by humans. In computational terms, theoretical linguistics strives to develop a causal model of how we \textit{learn}, \textit{parse} and \textit{generate} language. A central figure whose work helped define this goal is Noam Chomsky. Early in his career, Chomsky contributed to theoretical linguistics in two major ways that characterized his approach. First, he developed computational formalizations of basic linguistic concepts such as ``language'' and ``grammar'' as well as mathematical proofs on the complexity of different formal grammars \cite{chomsky1956threeModels, chomsky1957ss, ChomskyMiller1958, chomsky1959algebraic, chomsky1959formal, Chomsky1963FormalGrammars}. This in turn contributed to computer science by creating the field of formal language theory, advancing the theory of computation, and paving the way for rule-based symbolic models of language used in tasks such as machine translation. Second, Chomsky laid out the philosophical and scientific foundations for the development of linguistic models and theories, as well as their evaluation against empirical data \cite{chomsky1957ss, chomsky1975lslt, Chomsky1964CurrentIssues, chomsky1965aspects, Chomsky1966Cartesian}. His ideas and early works shaped theoretical linguistics, established the field of \textit{generative linguistics}, gave rise to the cognitive revolution, and set off the wave of symbolic Artificial Intelligence (AI) in mid-20th century. 

More recently, there has been a new wave of non-symbolic AI research, based on neural and statistical machine learning, access to large amounts of data, and increased computing power. Neural language models (LMs) have shown impressive linguistic capabilities, surpassing other AI models on many natural language processing tasks \cite{srivastava2023beyond,mahowald2023dissociating}. Some now question whether generative linguistics can continue to progress independently from LMs and generative AI research \cite{pater2019generative, potts2018case} or even its relevance as an approach at all \cite{piantadosi2023modern}, leading to heated debate \cite{bender2021dangers, baroni2022proper, markus2022deep, dijk2023large, kodner2023linguistics, katzir2023large, chomsky2023noam, lappin2024assessing}.

We argue that LMs reinforce some of the basic tenets of Chomsky’s approach to language, and that in turn, generative linguistics helps the evaluation and improvement of generative AI models. More specifically, we claim that: 1) LMs are formal generative models as intended originally in Chomsky’s work on formal language theory; 2) LMs can help develop a program for discovery procedures as defined by \citeA{chomsky1957ss}; and 3) LMs can be a major asset for Chomsky's minimalist approach to Universal Grammar and language acquisition. We expand on each of these points in the following sections. We agree with others who have argued that LMs are simply another tool in the toolkit of theoretical linguists and cognitive scientists \cite{futrellMahowald2025}. We also explain how Chomsky’s approach and generative linguistics contribute to research on LMs, providing the framework in which these models can be developed, evaluated, and improved to serve the scientific goals of the field.

\section{On Generative Models, Generative Capacity, and Levels of Adequacy}\label{sec:adequacy}

\begin{figure}
    \centering
    \includegraphics[width=\textwidth]{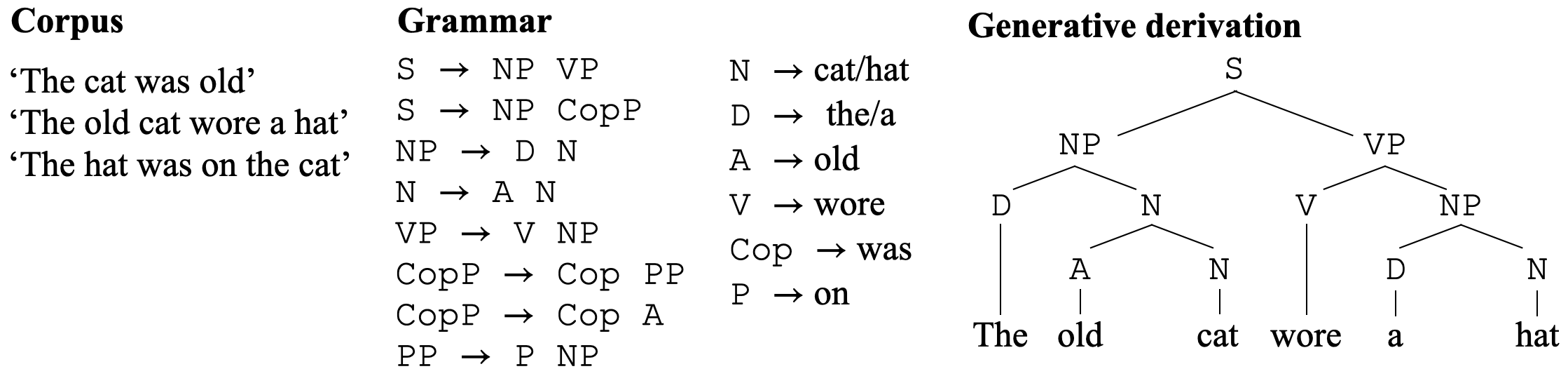}
    \caption{Example of a simple context-free generative grammar and derived sentence, based on a corpus}
    \label{fig:gram-ex}
\end{figure}

\citeA{chomsky1956threeModels, chomsky1957ss, chomsky1959formal} defined a \textit{language} formally as a possibly infinite set of finite-length sentences constructed out of a finite set of elements (i.e. words or subword units). He defined a \textit{(generative) grammar} as a "device" (i.e. a model) that generates all the grammatical sentences and none of the ungrammatical ones. This way, the task of finding the right grammar for a natural language such as English or French, was transformed into the task of finding the right computational device or model that generates the sentences of that language, see Figure 1. Since human languages share considerable structural similarities, it is reasonable to assume that the grammars that generate them are also similar structurally. Chomsky argued that linguistic theory should also provide a general grammar or class of devices that captures the patterns common to all human languages, and spells out the ways languages can be different from each other. For example in Figure 1, this general theory could specify how many latent categories are possible in a language or even what their properties must be -- verbs (\texttt{V}) must imply state changes or nouns (\texttt{N}) must be referents -- but how these categories are assigned to words would be language dependent. Later, this general grammar or device that captured universals of human languages as well as their range of variation was called Universal Grammar (UG).

\begin{figure}
    \centering
    \includegraphics[width=\textwidth]{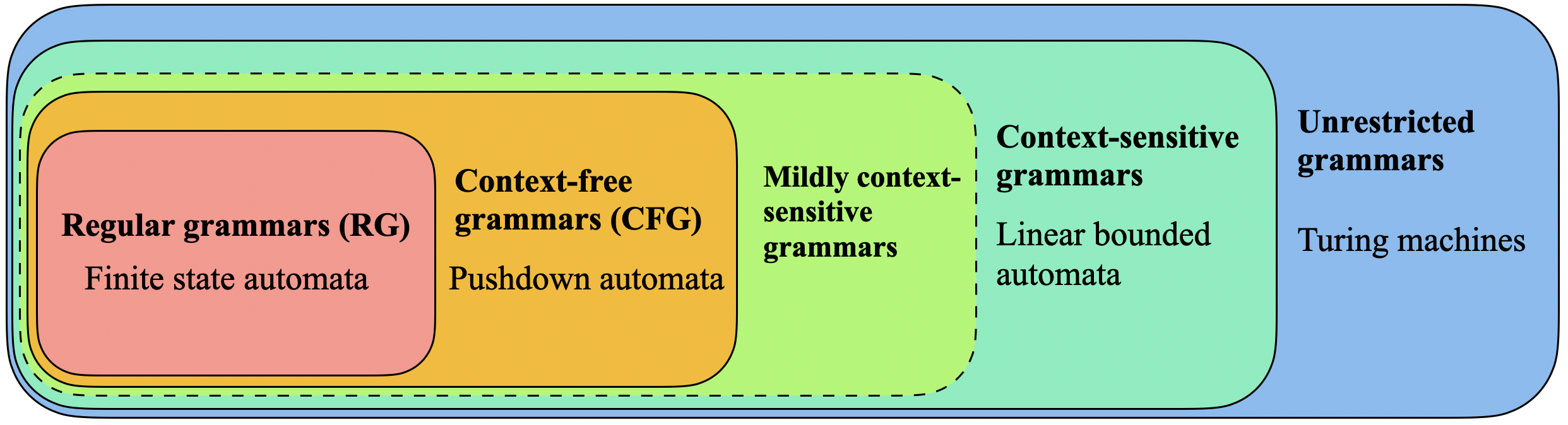}
    \caption{Chomsky-Schützenberger hierarchy}
    \label{fig:hierar}
\end{figure}

As part of his research on machine translation, Chomsky investigated common computational models of the time including finite state automata and push-down automata, as well as their corresponding grammatical descriptions namely regular grammars (RGs) and context-free grammars (CFGs) \cite{Chomsky1956MITreportAdequacy, Chomsky1956MITreportFSAs, chomsky1956threeModels, Chomsky1958MITreportPSGs, Chomsky1962MITreportCFGs}. Along with the mathematician Marcel-Paul Schützenberger, he established a hierarchy of increasing complexity among automata and formal grammars, later called the Chomsky-Schützenberger hierarchy (Figure \ref{fig:hierar}). Chomsky argued that neither RGs nor CFGs are adequate grammars for capturing the complexity of natural languages \cite{chomsky1956threeModels, ChomskySchutzenberger1963, chomsky1957ss, ChomskyMiller1963MathHandbook}. Therefore linguists should look at formalisms beyond CFGs. These arguments set off a productive line of research on different types of machines, formal grammars, their computational complexity, and their ability to generate natural language sentences \cite{JoshiEtal1975, Aho1968, Gazdar1988, BarHillel1964, GazdarEtal1985, JoshiEtal1991, PetersRitchie1973, TorenvelietTrautwein1995, Stabler2011, Michaelis2001}. This literature has converged on the conclusion that the computational complexity of natural languages falls in the ``mildly context-sensitive" region of the Chomsky-Schützenberger hierarchy.

More recently, there has been renewed interest in investigating new classes of machines, formal grammars, and languages. For example, \citeA{icard2020calibrating} studied probabilistic generative grammars, and showed that the probabilistic Chomsky-Schützenberger hierarchy has important differences, including the fact that probabilistic context-sensitive grammars are incomparable with the probabilistic context-free grammars. Similar investigations have been under way to compare neural generative LMs with the automata in the classical Chomsky-Schützenberger hierarchy. For example, \citeA{deletang2023} tested Recurrent Neural Networks (RNNs), Transformer-based LMs, and versions augmented with classical stack or tape memory such as stack-RNN and tape-RNN. They grouped their language recognition tasks based on the Chomsky-Schützenberger hierarchy to test whether different neural architectures are able to generalize to out-of distribution inputs. They showed that RNNs and Transformers models fail to generalize on non-regular language tasks while LMs augmented with structured memory successfully generalize in context-free and context-sensitive language tasks. It is too early to have a hierarchy or general theory for the computational complexity of different LMs but future investigations can provide a unified account for classical, probabilistic, and neural generative models.

Generative grammars developed by Chomsky have changed over the years, starting with CFGs that generated base form, plus rules that transformed them into the output s\cite{chomsky1957ss, chomsky1965aspects}, to binary-branching CFGs with a fixed X-bar schema and limited transformations \cite{Chomsky1981}, to using a binary set-forming operation called Merge \cite{chomsky1995minimalist, ChomskyEtal2023}. More recently and in collaboration with Matilde Marcolli and Robert Berwick, Chomsky has considered formalizations of his Minimalist syntax in terms of Hopf algebras and vector space representations \cite{marcolliChomskyBerwick2025}, potentially bringing Minimalist syntax closer to representations and formalizations used by LMs. While defining language formally and viewing grammars as computational models that generate them is an integral part of Chomsky's generative approach, the specific grammar or computational model is not. Any model, including LMs, can serve the goals of generative linguistics as long as it meets certain scientific criteria or conditions of adequacy.

Chomsky defined three levels of adequacy for grammars or models of language: 1. observational adequacy, 2. descriptive adequacy, and 3. explanatory adequacy \cite{Chomsky1964CurrentIssues, chomsky1965aspects}. An observationally adequate grammar or model ``presents the observed primary data correctly''. We can say that a grammar or model of language $L$ is observationally adequate if all the sentences it generates are in $L$ and it generates no sentences that are ungrammatical (not in $L$). More informally, a model's external behavior is identical to the external behavior of the speakers of $L$. Observational adequacy is also related to the concept of ``weak generative capacity", introduced by \citeA{chomsky1959formal}. The weak generative capacity of a grammar or model $G$ is the language $L$ that is generated by the grammar or model. Chomsky emphasized that evaluating models purely based on observational adequacy or their weak generative capacity is insufficient for scientific inquiry. A variety of models with different architecture and degrees of complexity can generate the same language $L$. To discover the model that best fits human knowledge of language, we need to go beyond the model's external behavior, and look at the internal properties of the model.

\begin{figure}
    \centering
    \includegraphics[width=0.95\columnwidth]{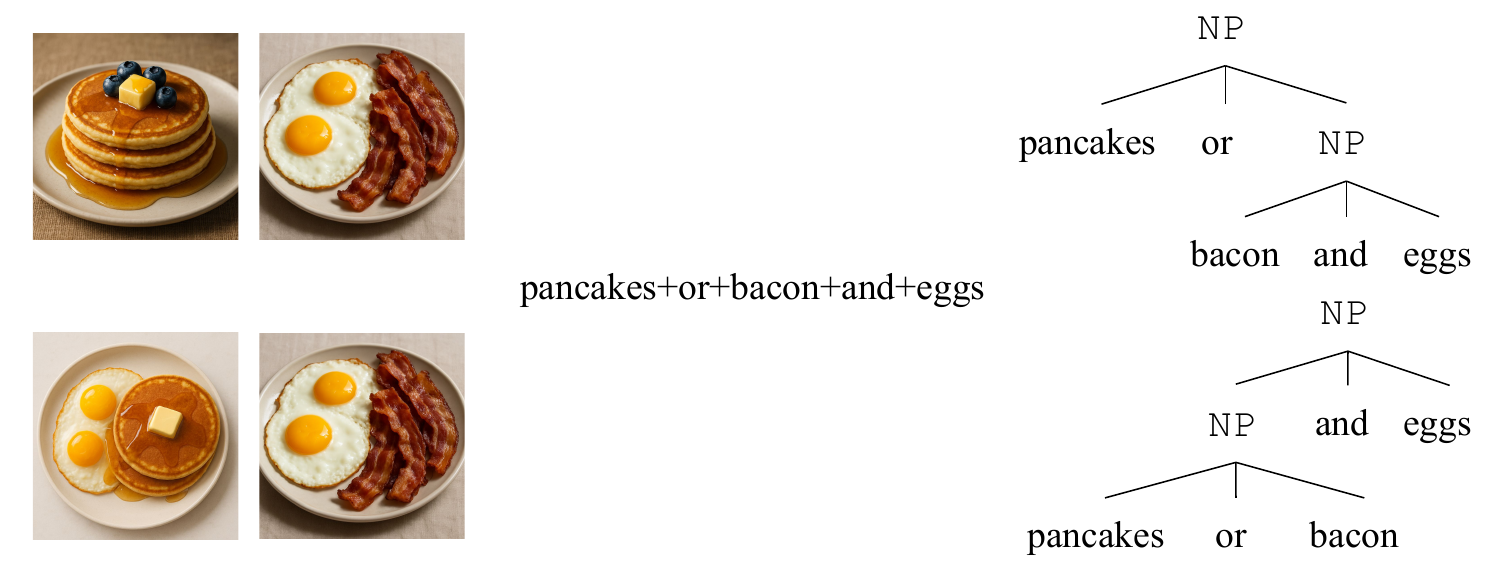}
    \caption{The structural ambiguity in the phrase ``pancakes or bacon and eggs" captured by a context free grammar (CFG) as two different hierarchical organization of the words: (pancakes or (bacon and eggs) vs. (pancakes or bacon) and eggs), but not captured by a Regular Grammar (RG) that simply concatenates words into the string ``pancakes+or+bacon+and+eggs".}
    \label{fig:pancakes}
\end{figure}

In addition to being observationally adequate, a ``descriptively adequate" grammar or model of language will  assign to each sentence the right structural description. In other words, it not only generates the sentences of the language, but additionally does so using an algorithm or process that matches what the human mind does at some level of abstraction. For example as speakers of English, we share the intuition that the phrase ``pancakes or bacon and eggs'' is ambiguous (Figure \ref{fig:pancakes}). It can refer to a choice between pancakes alone vs. bacon and eggs together, or pancakes and eggs vs. bacon and eggs \cite{RichCline2014}. The phrase itself can be generated from the set of words \{pancakes, bacon, eggs, or, and\} using a finite state automaton or a Regular Grammar using the string concatenation operation shown with plus +, or a push-down automaton and a constituency grammar like CFGs with the constituent formation operation shown with parentheses.

While both automata or grammars succeed in generating the string, they do so by assigning different structural descriptions. String concatenation can assign only one description in which each word follows the previous one like beads on a string. The constituent formation operation, however, can create the same string in two ways: once connecting bacon with eggs using "and", and another connecting bacon with pancakes using "or". These two constituent structures correspond to the two meanings we intuitively derive from the string. Therefore, the push-down automaton and corresponding CFG succeeds in capturing the structural ambiguity. In this small example, while both automata and associated grammars achieve observational adequacy by generating the string ``pancakes or bacon and eggs'', only the CFG achieves descriptive adequacy with respect to our intuition regarding the structural ambiguity of the phrase. Descriptive adequacy is also related to the notion of ``strong generative capacity", defined as the set of structural descriptions generated by the grammar.

Finally, in addition to achieving observational and descriptive adequacy, a grammar that reaches explanatory adequacy should explain why the language learner has arrived at this particular grammar as opposed to other possible grammars given the data available to them \cite{chomsky1965aspects, Chomsky1967innateIdeas}. In other words, a model of language with explanatory adequacy should learn from the amount and type of data available to children and match their comprehension and production milestones in linguistic development. Large LMs (LLMs) fall especially short in explanatory adequacy because they use orders of magnitude more data for learning than children \cite{Frank2023gap} and still under-perform in many basic tasks such as understanding negation \cite{truong-etal-2023, alhamoudEtal2025}, which emerges very early in child language development \cite{LiuJasbi2025, DeCarvalhoEtal2025, FeimanEtal2017}.

We can extend the notions of observational, descriptive, and explanatory adequacy to all aspects of cognition and external behavior using what we call ``the cognitive duck test''. In 1783, the French inventor Jacques de Vaucanson created a mechanical duck called ``Canard Digérateur'' (Digesting Duck), that could move its neck to muddle water with its bill, drink water, take food from someone's hand, swallow the food and excrete what appeared to be a digested version of the food \cite{riskin2003defecating}. The digesting duck likely inspired the colloquial duck test: ``If it looks like a duck, swims like a duck, and quacks like a duck, then it probably is a duck''. While the duck test itself is not a valid form of reasoning, the cognitive variant we define here can help with understanding levels of adequacy for models. If an automaton or model of a duck looks like a duck, walks like a duck, quacks like a duck, and more generally is indistinguishable from a duck behaviorally and externally, then it is an observationally adequate model of a duck. If it also has the internal structure and functions of a duck, then it is a descriptively adequate model of a duck. Finally, if the model also develops and evolves similar to what we know about the development and evolution of ducks, then the model is an explanatorily adequate model of a duck.

\section{On Procedures for Linguistic Theory Development}

 \begin{figure}
    \centering
    \includegraphics[height=4cm]{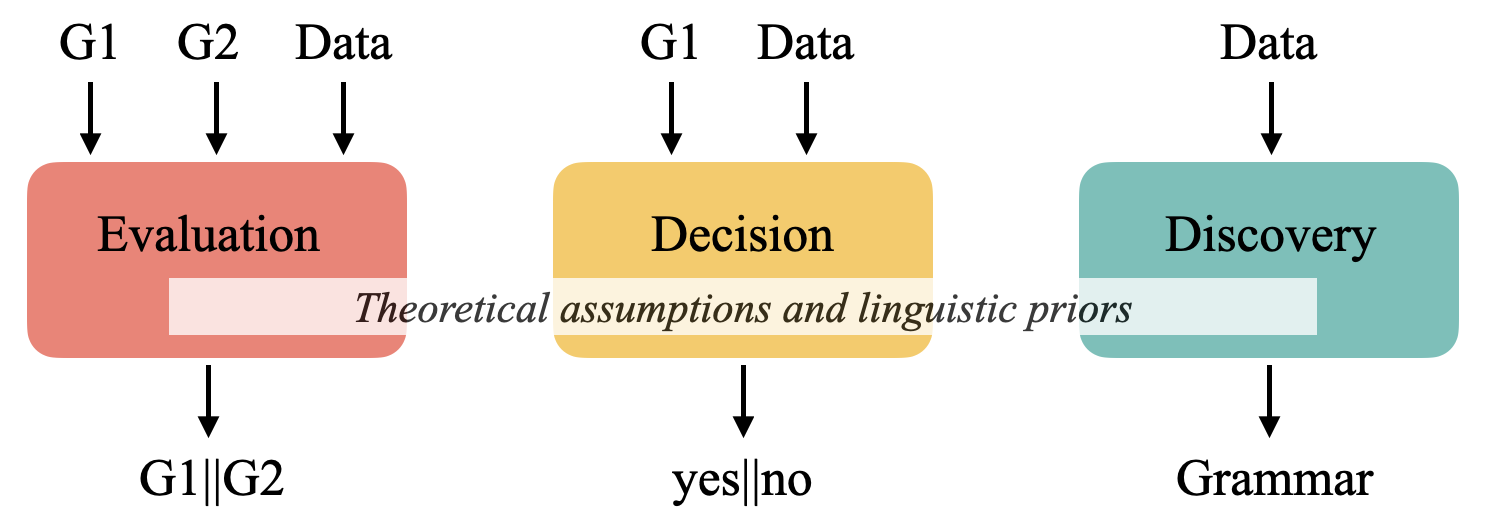}
    \caption{Procedures for linguistic theory development. Inspired by the original figure from \textit{Syntactic Structures}, \protect\citeNP{chomsky1957ss}.}
    \label{fig:procedures}
\end{figure}

\begin{figure}
    \centering
    \includegraphics[height=7cm]{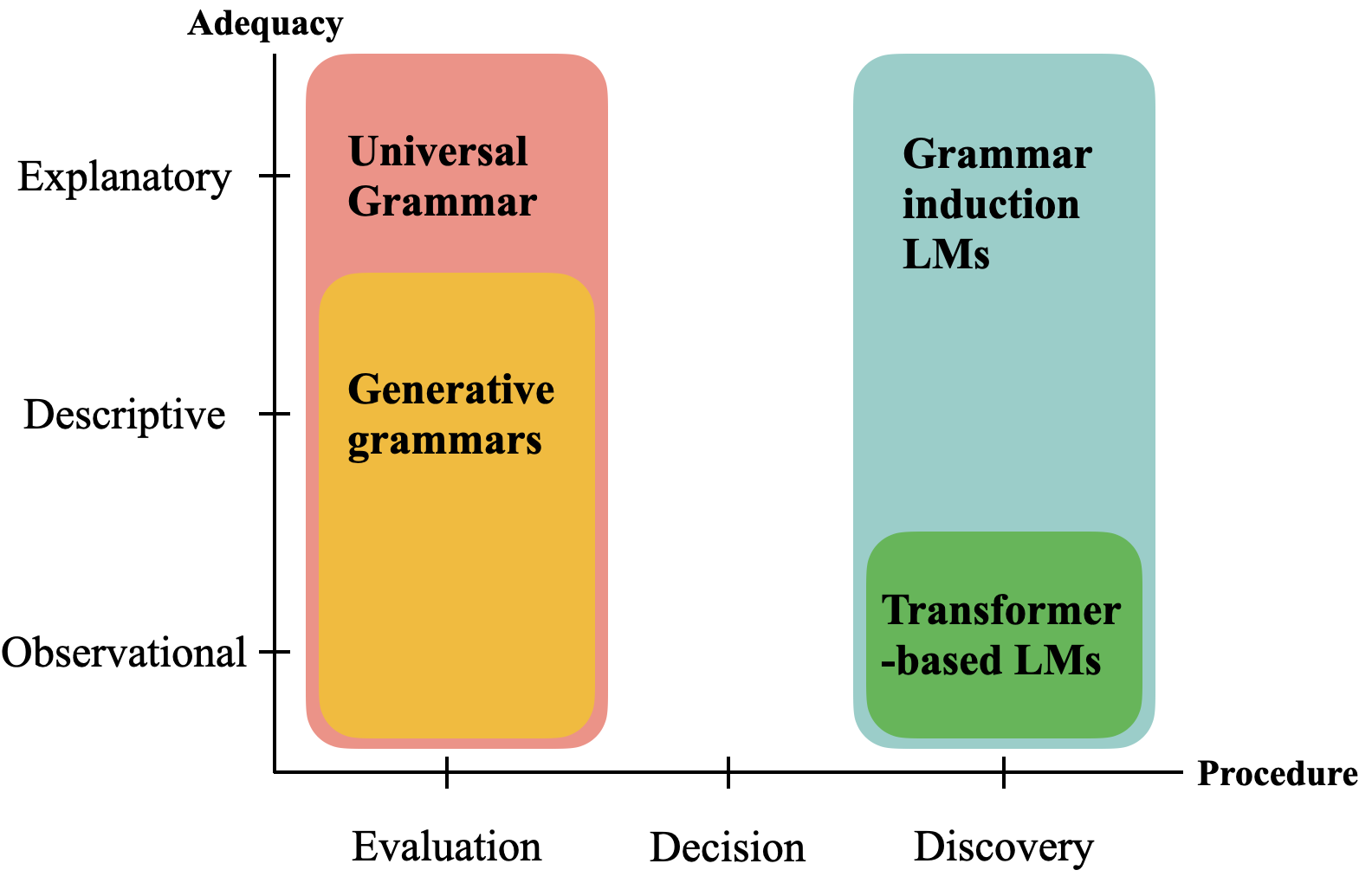}
    \caption{The relative positions of both linguistic and neural network based generative theories of language according to their level of theoretical adequacy and their procedural type.}
    \label{fig:comparative}
\end{figure}

All theories make assumptions about how languages are represented or learned, what differentiates them is the quantity and qualities of assumptions made. Chomsky \citeyear{chomsky1957ss} defined three types of theoretical procedures: discovery, decision, and evaluation procedures (Figure \ref{fig:procedures}). (1) A discovery procedure takes data and returns or \textit{discovers} the best or most likely grammar to have generated said data. (2) A decision procedure takes data and a grammar and \textit{decides} whether or not this grammar is the best or most likely to have generated said data. (3) An evaluation procedure is one which takes data and a set of possible grammars and \textit{evaluates} which of grammar from the set is a best or most likely candidate to have generated said data. These procedures are sets of theoretical assumptions that delimit the space of possible grammars under consideration. In all cases, a procedure could have either observational, descriptive or explanatory adequacy, depending on what theoretical assumptions it makes, as illustrated in Figure \ref{fig:comparative}. 

Generative linguistics has mostly relied on evaluation procedures, evaluating the descriptive adequacy of a predefined grammar over others. 
Discovery procedures require only language data as their input. They do not require making strong assumptions about linguistic structure by predefining grammars under study; Instead, they induce or \textit{learn} a complete grammar as their outcome. Discovery procedures may thus be particularly useful for offering explanations for how language may be acquired. However, coming up with successful ones was until recently nontrivial. When Chomsky wrote \textit{Syntactic structures} \citeyear{chomsky1957ss}, there were two discovery efforts. First, the American structuralist movement tried to provide a general method for determining the grammar of a language from a corpus using sets of heuristic rules across levels of linguistic representations \cite{harris1951methods}. Second, the recent application of Markov models to language, known as n-gram language models, provided the first computational discovery procedure \cite{shannon1951prediction}. N-gram models are defined by Equation \ref{eq:ngram}, where $n$ is a hyperparameter representing the word co-occurrence context size being considered and $w_{1}, ...,w_{n}$ are a sequence of words. Simply put, they return the normalized count of words in each context of length $n$ in a corpus.
\begin{equation} \label{eq:ngram}
  P_{n-gram}\left(w_n \mid w_{1}, ...,w_{n-1} \right)
\end{equation}
In response to the heuristic approach, Chomsky argued that such sets of rules would likely be too difficult to elaborate, leading to a ``maze of more and more complex analytic procedures''. He himself tried to develop an algorithmic discovery procedure in his earliest paper \cite{chomsky1953systems}, which did not get much traction due to its complexity. As for the computational approach, he argued that these n-gram models should be ruled out because they could not account for grammaticality judgments, making no distinction between grammaticality and likelihood; in other words they were only observationally adequate. For example,  ``Colorless green ideas sleep furiously'' is a grammatical sentence though nonsensical, making the chance of having seen such n-grams near impossible. Thus, the model cannot distinguish such a sentence from the just as unlikely yet ungrammatical one ``Furiously sleep ideas green colorless''.\footnote{Unless of course you've taken a linguistics course, in which case it is highly likely that you have seen these sentences before.} More recently, some have made similar arguments about LMs' inability to distinguish grammaticality from likelihood \cite{chomsky2023noam, kodner2023linguistics, katzir2023large}, ruling out these contemporary models as valid discovery procedures going beyond observational adequacy. However, these arguments are built on the assumption that LMs can only provide us with \textit{string} likelihood and not \textit{derivational} likelihood. In other words, They assume that models can only provide us with distributions over strings (like n-gram models) and not distributions over grammars or latent structures that derive those strings. We will argue that LMs can in principle learn grammars and that some decidedly do so, making a clear distinction between string likelihood and derivational likelihood, or grammaticality.

\begin{figure}
    \centering
    \includegraphics[width=\textwidth]{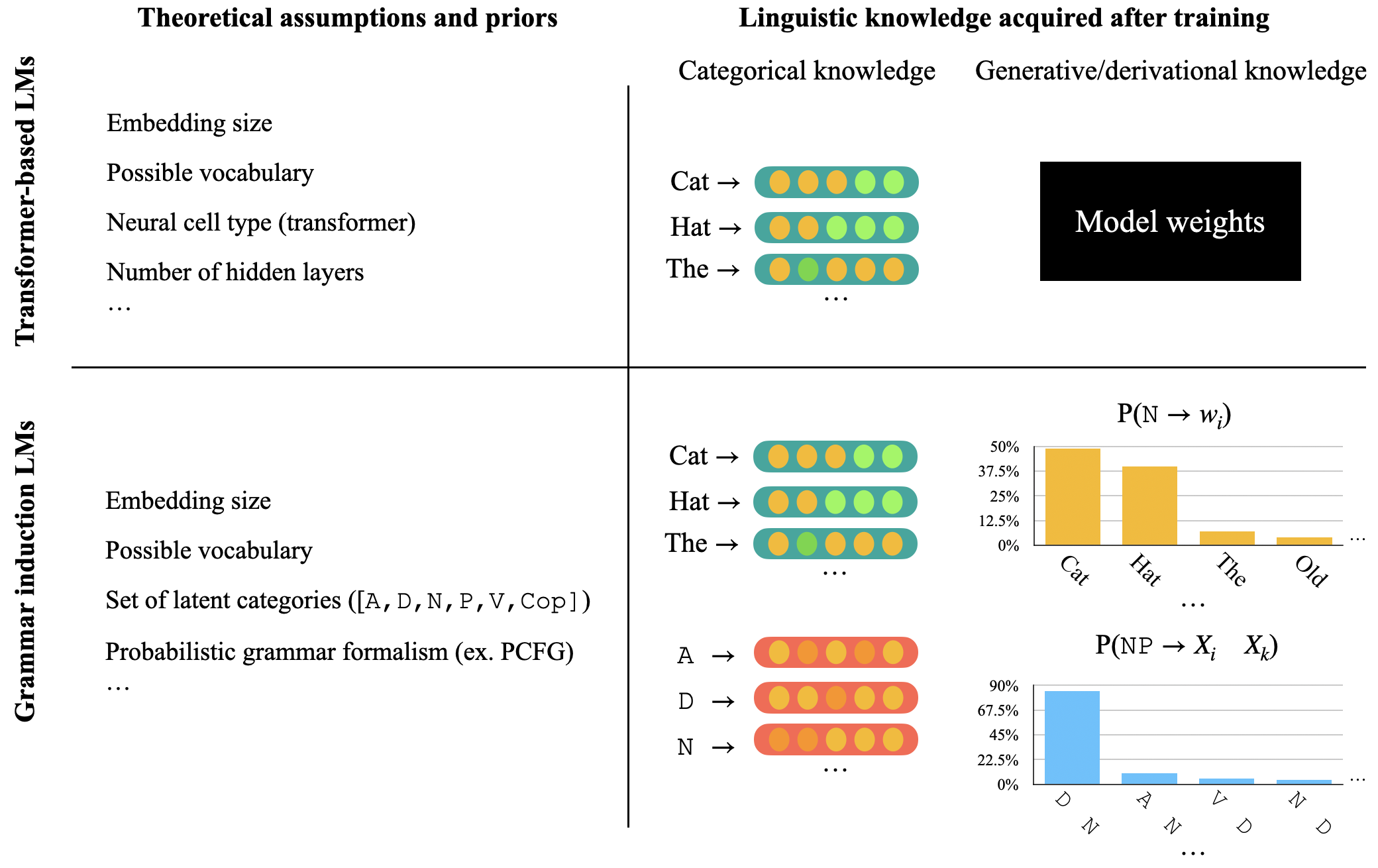}
    \caption{The differences between transformer-based LMs and grammar induction LMs in terms of their linguistic priors and the knowledge they acquire. Categorical knowledge is represented with distributed representations, or embeddings. In this example, for the grammar induction LM, we might predefine the distributional properties of each lexical category, but we could simplify our assumptions even more and have an arbitrary set of categories ([$\texttt{C}_1, ..., \texttt{C}_k$]), thus needing to learn their properties and embeddings from scratch. These two versions would represent two different hypotheses about UG, one where prior knowledge of linguistic categories is necessary and one where it is not.}
    \label{fig:LM-graminduct}
\end{figure}

LMs make very few assumptions about the nature of linguistic structure, making them promising discovery procedures. We may think of the untrained model architecture and hyperparameters as defining the theoretical assumptions under which a generative model for any language can be learned. While the trained model which has been fitted to specific data as akin to a specific language grammar or procedure output, see Figure \ref{fig:LM-graminduct}. Like n-gram models, LMs learn word co-occurrence statistics, but they additionally learn distributed representations, or embeddings, for words and neural model weights. The embeddings encode sets of shared linguistic features across words, while neural model weights encode how these features can be combined -- much like higher order grammatical rules. Thus, LMs can \textit{in principle} learn latent structures that could account for the distinction between unseen or unlikely sentences versus \textit{impossible} ungrammatical sentences.

String likelihood under an LM is the probability of sequentially sampling a set of words from the final layer of model weights. In practice this value is calculated using the Markov assumption: the probability of the next word in a sequence is only dependent on the current state of the model's final layer. Thus, string likelihood from an LM has the same flaws as string likelihood from an n-gram model, also known as a Markov model. However, derivational likelihood under an LM is the probability that a string be derived given all relevant model weights and states, without this assumption. It encodes the model's latent grammatical knowledge about a sentence. The problem is not that this value doesn't exist but rather that we do not always have a clear way of extracting it, LMs often being referred to as black boxes. 

Thus, a related and fairer line of argumentation against LMs usefulness for linguistic theory is that they do not give us a description of or explanation for our linguistic knowledge \cite{kodner2023linguistics, chomsky2023noam}. Though methods such as principal component analysis, probing methodologies \cite{hewitt2019structural}, or  prompting strategies currently exist for retrieving \textit{some} of the internal knowledge of transformer-based LMs, like the GPT models or encoder models such as BERT \cite{devlin-etal-2019-bert, radford2019language}, these methods are predicated on strong linguistic inductive biases that indicate ``what to look for'' \cite{kodner2023linguistics}. We cannot yet extract the grammars these LMs learn from a set of more general assumptions about generative grammar properties. If methods to do so were developed,  then these models could aspire to descriptive adequacy. Though transformer-based LM architectures are currently the most prominent, they are not the only neural networks that ``learn language'', an alternative are neural grammar induction models (grammar induction LMs)  \cite{kim-etal-2019-compound, drozdov-etal-2019-unsupervised-latent, zhu-etal-2020-return}. These models are often overlooked, likely because implementing them and studying them requires a joint expertise in both machine learning and formal language theory, a combination of skills less seen in recent years.

\textit{Grammar induction} refers to the computational task of taking a corpus of grammatical sentences and inducing a probabilistic generative grammar, or set of categories and rules, that maximizes the probability of deriving the corpus under the constraints of the induction program. Grammar induction LMs learn distributions over explicit sets of syntactic rules, or \textit{descriptions} of the data, making derivational likelihood easily accessible, see Figure 4. 
They can learn to group words into salient syntactic categories and to combine these categories into generative rules for forming constituents, giving equivalently descriptive grammars to existing generative grammars. They do so with very few theoretical assumptions, though more than transformer-based LMs require. Grammar induction LMs represent the best of both worlds: (1) they can be used as discovery procedures, making few prior assumptions about the structure of language, much like other LMs, yet (2) like generative linguistic theories they strive towards descriptive and even explanatory adequacy. They can model not only the ``end result'' or ``adult'' grammars induced, but also the language acquisition process as a whole and have already been used to do so \cite{portelance2024reframing}. Thus, grammar induction LMs, as discovery procedures, represent a set of AI models that could contribute to theoretical linguistics, especially theories of language learnability that are both descriptively and explanatorily adequate.

\section{On Learnability, Language Acquisition, and Universal Grammar}

An important goal of linguistic theory is to explain how humans learn natural languages and how human learning shapes language structure. Chomsky's formalization of linguistic concepts and the development of the Chomsky-Schutzenberger hierarchy paved the way for the formalization of language learning and the development of \textit{formal learning theory} \cite{Osherson1984, osherson1997formal, oshersonEtal1990}. Language learnability became the formal and computational study of how any agent, human or machine, can learn a language, given certain assumptions about the available data to the learner and the learning mechanism used by the agent.

In a very influential paper, \citeA{gold1967language} defined a learning paradigm called ``language identification in the limit'' in which a learner must try to identify a grammar $G$ for an arbitrary formal language $L$ from some language class $\mathcal{C}$ of the Chomsky-Schutzenberger hierarchy. In this paradigm, the learner is presented with a sequence of strings $s_1, ..., s_t$ of $L$, and at each time $t$ guesses a grammar $G_t$ which explains all the sequences it has seen up to that point. A \textit{class} of formal languages $\mathcal{C}$ was defined as \textit{learnable} if and only if, under any possible presentation of the data, the learner could guess in finite steps the correct grammar. He tested two methods of presenting data to the learner: one without an informant (positive evidence only) and one with an informant labeling each string as grammatical or ungrammatical (positive and negative evidence). He proved that with both positive and negative evidence, up to context-sensitive languages are learnable in the Chomsky Hierarchy, while with positive evidence alone, not even the class of regular languages is learnable. With respect to human language acquisition -- since children clearly do learn their language -- \citeA{gold1967language}'s results suggested the following hypotheses: 1. human languages are a subclass of formal language classes in the Chomsky-Schutzenberger hierarchy and human language learning is innately restricted to this subclass; 2. children rely on some form of negative evidence as well as positive evidence; or 3. the language identification in the limit paradigm has unrealistic assumptions that need to be relaxed or discarded. For example, children may not need to learn from every possible presentation of data and child-directed speech may be a special presentation of linguistic evidence; or children may not need to learn the exact grammar $G$, but rather learn a grammar that is ``approximately'' $G$ or close enough.

Early theories of language learning in generative linguistics took hypothesis 1 above seriously and used the theory of Universal Grammar (UG) to address the issue of learnability by positing innate linguistic knowledge, such that a learner only searches through a subclass of formally possible languages and grammars. \citeA{chomsky1965aspects} proposed that UG contains an enumeration of possible natural language sentences, grammars, and structural descriptions that those grammars assign to sentences. It also contains an evaluation metric that assigns a value or score to each grammar, comparing grammars and selecting one that best matches the observed data. Similar to Chomsky's approach to linguistic theory, he addressed learnability by formulating language learning as an evaluation procedure with a restricted class of innate grammars. Over the years, there have been several proposals regarding the content of UG based on the prevalent syntactic theory at the time. For example, \citeA{HamburgerWexler1973, HamburgerWexler1975, WexlerHamburger1973} suggested that UG contains a universal base of context-free rules and a set of transformations that vary between languages. They investigated the learnability of such systems using Gold's paradigm of identification in the limit and reported some positive learnability results. Following the shift in syntactic theory from transformations to universal principles and language-specific parameters \cite{Chomsky1980, Chomsky1981}, several researchers proposed learning models of UG as evaluation procedures with a fixed number of parameters specific to language learning. The role of data was to \textit{trigger} or set particular binary parameter values \citeA{GibsonWexler1994, sakasFodor2012} or \textit{punish}/\textit{reward} them in a gradient manner \citeA{Yang1999, Yang2000}.

More recently, generative linguistics has also engaged with issues related to hypotheses 2 and 3 suggested by Gold's results, and more generally, has explored factors other than innate knowledge that can address the problem of language learnability. \citeA{Chomsky2007, Chomsky2005} explained that language learning relies on three factors: 1. Genetic endowment for language (UG), 2. Experience and primary linguistic data, and 3. General principles of learning, processing, communication, efficient computation, etc. Since language has evolved rather quickly on the evolutionary time-scale, its innate factor (i.e. Universal Grammar) has had little time to evolve, and therefore must be simple and minimal, offering an optimal solution to the problem of mapping concepts and thought to linguistic sounds and gestures \cite{HauserChomskyFitch2002, berwick2016only}. In his \textit{Minimalist Program} \citeyear{chomsky2001minimalist}, he suggests that to discover the genetic endowment for language (factor 1), researchers could discover what aspects of human language are explainable using the data available for learning (factor 2) and general principles of learning, processing, communication, etc. (factor 3). What remains unexplained by factors 2 and 3, will be a good candidate for something contributed by the first factor, namely genetic endowment for language.

\begin{quote}
Throughout the modern history of generative grammar, the problem of determining the character of the language faculty has been approached ``from top down": How much must be attributed to Universal Grammar to account for language acquisition? The Minimalist Program seeks to approach the problem ``from bottom up": How little can be attributed to Universal Grammar while still accounting for the variety of [internal language knowledge] attained, relying on third factor principles? The two approaches should, of course, converge, and should interact in the course of pursuing a common goal.
\end{quote}

LMs can be a major asset in the Minimalist Program's ``bottom up" approach because they generally assume little to no domain-specific knowledge, and rely mainly on statistical properties inherent in data as well as general mechanisms of learning and processing. Grammar induction LMs may be even more helpful due to their transparency and the interpretability of induced linguistic generalizations and representations. These LMs can computationally implement different hypotheses regarding the roles of factors 2 and 3 explained above. Their systematic failures provide us with hypotheses for what the role of factor 1 (genetic endowment) could be. In turn, the discovery of innate factors that help language learning in humans, can also make language learning in LMs and more generally AI systems more efficient.

\section{Final Discussion} 

We started this paper by highlighting the scientific goals of theoretical linguistics: understanding what language is and why it is this way. One of the leading schools of thought in attaining these goals since mid 20th century is generative linguistics, Noam Chomsky's works having laid its computational and theoretical foundations. More recently, generative AI and Language Models (LMs), primarily developed forengineering purposes, have shown impressive performance on linguistic tasks. We therefore asked what role can LMs play in helping generative linguistics achieve its scientific goals and how can generative linguistics contribute to the development of better performing LMs? We argued that LMs are compatible with Chomsky's generative approach and can contribute to each other in at least 3 major ways.

First, LMs fit Chomsky's original formalization of grammars as automata that generate sentences of a language. Despite considerable differences between LMs and the symbolic automata of generative grammars, they can strive towards the same criteria of adequacy established by Chomsky, namely observational, descriptive, and explanatory adequacy. In line with their early engineering goals, LMs have had considerable success establishing observational adequacy so far. But to contribute to scientific disciplines such as cognitive science and linguistics, they need to make advances in descriptive and explanatory adequacy as well (i.e. human-like linguistic knowledge, language processing, and learning). We are hopeful that striving towards descriptive and explanatory adequacy, LMs may improve their efficiency in learning and require smaller amounts of data or improve their performance on high level tasks like logical reasoning.

Second, Chomsky discussed three ways linguistic theory can build and evaluate grammars or more generally models of language; by providing discovery procedures, decision procedures, and evaluation procedures. We argued that LMs represent effective discovery procedures, in particular grammar induction LMs, as they may reach both descriptive and explanatory adequacy. Discovery procedures may be necessary to answer important fundamental questions about the form of grammars, because unlike other procedures they do not presuppose defined grammars, they learn them. Models may thus help progress inquiries such as: what are the fundamental categories of language?  Do we need movement or transformation-based operations or are formalisms with lexicalized, feature-based, or cross-category style rules enough?

Third, early research on the theory of Universal Grammar had the goal of discovering which characteristics of human language are likely innate. More recently, Chomsky's Minimalist Program proposes to do so by explaining away properties of language that can be accounted for by what he called third factors such as general cognitive learning mechanisms, principles of efficient communication and processing, etc. What remains and cannot be explained by third factors and the data available for learning, will be a good candidate for what the genetic component of language learning and processing may be. Since LMs typically use no language-specific knowledge prior to training beyond a base vocabulary, and rely on third factor elements and available data for learning, they are valuable computational tools for the Minimalist Program's goals. In turn, discovering innate components that aid human language acquisition will prove invaluable to generative AI research in developing models that are more data efficient and more successful at language learning.

In highlighting the compatibilities between generative AI and generative linguistics, we do not mean to deny the existence of challenges and inconsistencies between these approaches; nor do we wish to suggest they are compatible only in the three ways we have discussed. We simply wish to convey that beyond simple collaboration between two different approaches to language, generative AI and generative linguistics have deep and foundational connections and many shared goals. As a final promise, one of the most interesting ways in which AI models may influence the future of theoretical linguistics is by introducing the effects of multimodal or multichannel input to language learning research.  LMs can take into account not only corpora of sentences, but also other data modalities such as grounding in images and eventually sound and video -- this is true of grammar induction LMs \cite{shi-etal-2019-visually, zhao-titov-2020-visually, jin-schuler-2020-grounded, wan2022unsupervised, portelance2024reframing}. Such models along with proposals for the development of theories of a multimodal language theories \cite{perniss2018we, dressman2019multimodality, cohn2024multimodal} may offer a more complete picture of our true lived-experience of language processing and learning -- helping us move towards our scientific goals.

\newpage

\bibliographystyle{apacite}

\end{document}